\documentclass{article}
\usepackage{amsmath, bm, amsthm}
\usepackage{graphicx}
\usepackage{standalone}
\usepackage{svg}
\usepackage{url}
\usepackage{enumitem}
\usepackage{xfrac}
\usepackage{xcolor}
\usepackage{xcite}
\usepackage{amsfonts}
\usepackage{multicol}
\usepackage{standalone}
\usepackage{pifont}

\DeclareMathOperator*{\argmax}{argmax} 
\usepackage[square, numbers]{natbib}
\usepackage{appendix}

\usepackage[final]{neurips2021/neurips_2021}
\usepackage[utf8]{inputenc}

\title{Kernel Identification Through Transformers}
\author{
Fergus Simpson \\
Secondmind \\
Cambridge, UK \\
\texttt{fergus@secondmind.ai}
\And
Ian Davies \thanks{Work undertaken while at Secondmind} \\
InstaDeep \\
London, UK \\
\And
Vidhi Lalchand \\
University of Cambridge \\
Cambridge, UK \\
\And
Alessandro Vullo \\
Secondmind \\
Cambridge, UK \\
\And
Nicolas Durrande \\
Secondmind \\
Cambridge, UK \\
\And
Carl Rasmussen \\
University of Cambridge \\
Cambridge, UK
}

\begin{document}

\maketitle
\setcounter{footnote}{0}
\begin{abstract}

Kernel selection plays a central role in determining the performance of Gaussian Process (GP) models, as the chosen kernel determines both the inductive biases and prior support of functions under the GP prior. This work addresses the challenge of constructing custom kernel functions for high-dimensional GP regression models.
Drawing inspiration from recent progress in deep learning, we introduce a novel approach named \textit{KITT: Kernel Identification Through Transformers}.
KITT exploits a transformer-based architecture to generate kernel recommendations in under 0.1 seconds, which is several orders of magnitude faster than conventional kernel search algorithms. We train our model using synthetic data generated from priors over a vocabulary of known kernels.
By exploiting the nature of the self-attention mechanism, KITT is able to process datasets with inputs of arbitrary dimension.
We demonstrate that kernels chosen by KITT yield strong performance over a diverse collection of regression benchmarks. \end{abstract}

\section{Introduction}
In recent years deep parametric models have become a prominent class of model for supervised learning and have delivered impressive empirical performance over a wide range of tasks. An important limitation, however, is that in their conventional form deep models do not provide prediction uncertainty. While their Bayesian counterparts try to achieve this, they require significant modifications to the training procedure and are computationally expensive. Uncertainty quantification in deep models is widely considered to be an open problem, the large array of research proposing alternative Bayesian neural networks underscores this \citep{gal2016dropout, hernandez2015probabilistic,lakshminarayanan2017simple}.

On the other hand, kernel driven methods within the Bayesian framework like Gaussian processes (GPs) account for prediction uncertainty by design.
While GPs provide a flexible framework for inferring distributions over functions, the inductive biases are controlled by the kernel function\footnote{also called covariance function or covariance kernel}. A well chosen kernel will typically yield dramatically better performance than a poorly chosen one.

How should we learn expressive kernels for high-dimensional tasks? This has frequently been highlighted as a central question for the continued relevance of GP methods \citep{gretton2017new}.
This work uses representations generated by a deep neural network to identify suitably expressive kernels for high-dimensional GP regression tasks. Kernel recommendation is performed by a decoder with access to a large vocabulary of primitive kernels and products of primitive kernels. The decoder maps an encoded representation of a dataset to a kernel that can be used to model it. The representation is attained by encoding the dataset, treated as a sequence of (input, output) pairs $\mathcal{D} = \{\bm{x}_{i},y_{i}\}_{i=1}^{N}$, utilising the permutation-equivariant nature of self-attention networks.

By training KITT with a sufficiently rich vocabulary of kernels, it can predict suitable kernels for a diverse array of real datasets. This work presents the following novel contributions:

\begin{itemize}
\itemsep0em
    \item Inspired by the successes of image captioning networks, we develop a novel framework named KITT for amortised kernel search. KITT takes raw datasets for predictive modelling as input and proposes kernels composed from a large vocabulary of kernel functions. 
    \item KITT's architecture introduces two novel features: it is entirely agnostic to the length and dimensionality of the data we wish to perform inference on, and it offers double permutation invariance (this ensures its outputs are invariant to permutations in either input dimensions or data points).
    \item We show that KITT can deliver kernel predictions in under 0.1 seconds.
    \item We introduce a novel variant of the linear kernel which forms a key component of KITT's vocabulary. 
    \item We demonstrate that the kernels identified by KITT offer strong performance against other baselines which deal with kernel engineering in the context of GPs.
\end{itemize}

\section{Background}
\label{bg}

This section offers a brief review of the two topics which are central to this work, namely Gaussian Processes and Transformers.

\paragraph{Gaussian Processes.} GPs offer a highly versatile framework for predictive modelling \citep{rasmussen2006} with generalisation properties controlled by a kernel function parameterised by hyperparameters. The functional form of the kernel governs the global attributes of the supported functions, such as smoothness and periodicity.
However, a suitable kernel function is unknown \emph{a priori} and the choice of the kernel function is a fundamental model selection problem. Once a kernel has been chosen, training conventionally proceeds by learning a point estimate of the hyperparameters that maximise the GP log marginal likelihood $\bm{\theta}^{*} = \argmax_{\bm{\theta}}\log p(\bm{y}|\bm{\theta})$. The marginal likelihood is available in closed-form for models with Gaussian likelihoods. Below we briefly summarise the standard GP framework. 

GPs are distributions over functions from which one can sample realised function values for given inputs. Concretely, for observations $X=\lbrace\bm{x}_i\rbrace_{i=1}^N$, and positive definite kernel function $k_\theta(\cdot,\cdot)$, with hyperparameters $\bm{\theta}$, $f(\cdot) \sim \mathcal{GP}(\bm{0}, k_{\theta})$. Typically, we observe noisy realisations of the latent function which are corrupted with Gaussian noise, $y_{i} = f(\bm{x}_{i}) + \epsilon_{i}$, $\epsilon_{i} \sim \mathcal{N}(0, \sigma_{n}^{2})$, and infer the kernel hyperparameters through maximising the likelihood of the model.

The GP marginal likelihood objective, $p(\bm{y}|\bm{\theta})$, is obtained by marginalising the likelihood $\bm{y}| \bm{f} \sim \mathcal{N}(\bm{f}, \sigma^{2}_{n}\mathbb{I})$ over the prior $\bm{f}| X, \bm{\theta} \sim \mathcal{N}(\bm{0}, K_{\theta})$,
\begin{equation}
p(\bm{y}|\bm{\theta}) = \int p(\bm{y}|\bm{f})p(\bm{f}|\bm{\theta})\ d\bm{f} = \mathcal{N}(\bm{0}, K_{\bm{\theta}} + \sigma_{n}^{2}\mathbb{I}) \, ,
\end{equation}
 where $\bm{f}=f(X)$ denotes a vector of realised function values, and $K_{\theta}$ denotes the $N \times N$ covariance matrix corresponding to evaluations of the covariance function at the $N$ training inputs, $(K_{\theta})_{i,j} = k_{\theta}(\bm{x}_{i}, \bm{x}_{j})$.

A long standing question is how best to select an appropriate kernel function for a given task.
One approach is to search over a discrete space of kernels, defined by combining a selection of primitive kernels with a predefined grammar \citep{bach2009high, duvenaud2014automatic}.
Typically, a greedy search is performed to identify the kernel offering the best representation of the data. Ideally, the quality of the kernel is quantified by the Bayesian model evidence, which can be computed by marginalising the marginal likelihood over the hyperparameters. However, since the integral is challenging to compute, each kernel's suitability is instead usually determined via a proxy for the model evidence, such as the Bayesian Information Criterion (BIC) \cite{schwarz1978estimating}. 

For a principled, Bayesian approach to kernel design,  instead of selecting a single kernel, one ought to consider multiple candidates.  In other words, it is desirable to marginalise over the space of kernels, not only over the space of functions for a single kernel. 
\begin{equation}
p(\bm{y}|\mathcal{D}) = \sum_i p(\bm{y}| K_i, \mathcal{D}) p(\mathcal{D} | K_i) p(K_i) \, .
\end{equation}
This yields a rich  posterior distribution comprised of a mixture of Gaussians. 
A conventional kernel search makes three key approximations. First, that contributions from all but the single chosen kernel $K^*$ can be neglected; second, that contributions from all but the maximum likelihood hyperparameters $\theta^*$ can be neglected; third, that the proxy for the model evidence is a reliable one.
There are several regimes, for example where the data is sparse or noisy, where all three of these assumptions do not hold. We shall aim to improve upon all three of these issues.

\paragraph{Transformers}
Transformers are a form of deep neural network which rely upon the attention mechanism \citep{xu2015show} to capture global context.
While they were originally proposed to tackle machine translation tasks \citep{vaswani2017attention}, they have rapidly attained state-of-the-art performance in a number of other areas of machine learning \citep{jumper2020high,parisotto2021efficient,radford2021learning}. Of particular relevance to this work, they have been successfully applied to image captioning \cite{cornia2020meshed, yu2019multimodal}, a task which involves summarising the key characteristics of rich data in a grammatical form.  This has a striking parallel to the challenge of selecting an appropriate kernel, especially since a form of grammar can be used to construct a broad selection of GP kernels.

The self-attention mechanism of transformers naturally lends itself to the permutation-invariant setting, as demonstrated by \citet{zaheer2017deep}.
This invariance to permutations in the ordering of inputs has been exploited to create Set Transformers, introduced in \citet{lee2019set}. The AHGP model \cite{liu2020task} uses a variant of the Set Transformer to infer the hyperparameters of the spectral mixture kernel.

Like all deep networks, transformers thrive when presented with an abundance of training data. Fortunately, for the task we have at hand, the training set is unlimited in size as we may sample training data from GPs with known kernels.

As stressed by \citet{liu2020task}, selecting an architecture which reflects the appropriate invariances is vital. The AHGP model is designed to be invariant to permutations in the ordering of the datapoints. We go one step further, and introduce a model which is doubly permutation invariant: its output is also invariant to permutations in the input dimension (whereas AHGP is \emph{equivariant} to the input dimensions).

\section{KITT}
\label{kitt}

In this section we describe and motivate KITT, a network which takes as inputs a set of datapoints $\{\bm{x_i}, y_i\}$, and outputs a kernel recommendation in the form of a `caption', by utilising a kernel grammar. The code is available at \url{https://github.com/frgsimpson/kitt}.

\paragraph{Kernel Grammar and Vocabulary:} Throughout the GP literature, the most commonly used kernels  belong to a limited set of primitive functions. In this work, we utilise eight primitive kernels: the squared exponential; periodic;  white noise; three variants of the Matern kernel $\left(\frac{1}{2}, \frac{3}{2}, \frac{5}{2}\right)$; the cosine kernel, and our novel variant of the linear kernel. This list comprises six stationary isotropic kernels,  one stationary anisotropic kernel (cosine), and one non-stationary kernel (linear).  

From this small set of primitive kernels, we wish to construct a larger array of more expressive kernels. This can be achieved by leveraging the closure properties of kernel functions \citep{smola1998learning}. Permitted operations include addition, multiplication, convolution, composition, and affine transformations. For simplicity, this work shall only consider two operators: addition and multiplication.

\begin{figure*}[t]
\centering
\includegraphics[width=\textwidth]{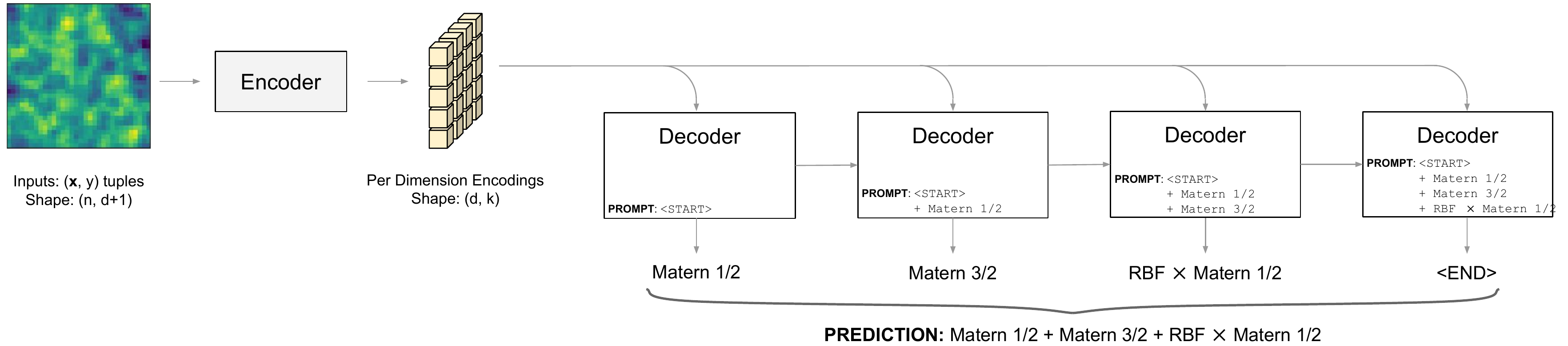}
\caption{The architecture for KITT, partly motivated by image captioning networks, which also act to transform a rich dataset into a grammatical expression of its contents. 
}
\label{fig:architecture}
\end{figure*}

Whilst this grammar of addition and multiplication appears superficially simple, it has some idiosyncrasies that would make it challenging for a network to learn. For example, if a network is unaware that multiplication is commutative, it would have to learn to recognise  $k_{1}*k_{2}$ and $k_{2}*k_{1}$ separately, for all combinations of $k_{1}$ and $k_{2}$. It would also need to learn that multiplying a noise kernel with another stationary kernel yields another noise kernel. Encoding this information \emph{a priori} greatly facilitates the learning process. To achieve this, we enlarge the vocabulary by defining product kernels as single `words', rather than incorporating products as part of the grammar (the full list is provided in the Appendix). This allows us to exclude redundant combinations from consideration. For example, only a single token is used to represent both $k_{2}*k_{1}$ and $k_{1}*k_{2}$. While this  enlarges the vocabulary, we can be confident that the network will be able to cope, since there will still be far fewer `words'  than can be found in natural language tasks where transformers are known to excel. We use products of two primitive kernels as part of the base vocabulary and found that incorporating higher order products do not significantly change performance. 

A further advantage of defining product kernels at the vocabulary level is that we need no longer include operators inside the vocabulary. The multiplication is already baked into the expanded set of kernels, while the addition takes place implicitly, much like the white space between words in a natural language task.
This precludes the construction of nested structures of operators, which would allow for even richer kernels. However, since this work represents a first attempt at performing a kernel search with a neural network, we choose to keep the captioning task to be a relatively simple one, and leave more complex grammatical compositions as an opportunity for future exploration. Even with this relatively simple grammar, if we permit a caption of four \textit{words} from a base vocabulary of 32, we are effectively searching across a space of around $36\,000$ kernels.

\paragraph{Priors.} While the priors we impose upon the hyperparameters will have some impact during optimisation, it is their influence upon the generation of KITT's training data that is of central importance to this work. Random samples are drawn from the hyperparameter priors $p(\theta)$ and input locations  $x \sim U(-2.5, 2.5)$ before each random sample of $y$ is generated. The variances and lengthscale parameters of all kernels (including product kernels) are assigned lognormal priors, such that $\log \theta \sim  \mathcal{N}(0, 1)$. 

The cosine kernel is unique among KITT's vocabulary, in that it is inherently anisotropic, which can be important if a preferred direction exists within the data. Samples drawn from the kernel manifest as plane waves which propagate along a characteristic direction of the kernel. If we were to impose a lognormal prior (or any other positively constrained prior) on its lengthscales, this would restrict the direction of the kernel to a small fraction $2^{1-D}$ 
of its permitted parameter space. We therefore adopt a different approach in this case, assigning $\ell \sim \mathrm{Cauchy}(0, 5)$ for the lengthscales.

\paragraph{Training.} 
KITT is trained entirely on synthetic data. Each training example is generated from a randomly selected kernel, with a randomly drawn set of hyperparameters. This data could be generated on the fly, but since this can be a computationally expensive process, we generated a training set in advance which comprised of $200,000$ labeled examples. While the model can be constructed for an arbitrary number of inputs points and input dimensions, during training we restrict ourselves to the case of 4 input dimensions and 64 input points per sample. The loss function corresponds to $-\log p(k | \mathcal{D})$, the negative log probability the network assigned to the correct term in the vocabulary. The Adam optimiser was used with an initial learning rate of $10^{-4}$, and a decay schedule with a decay rate of $0.1$ every $50,000$ iterations.
Due to the relatively noisy nature of the classification task, a relatively large batch size of 128 was found to be beneficial.
The vocabulary included product kernels of at most two terms in addition to the primitive kernels, yielding a final vocabulary of size 34.

\paragraph{Heteroscedastic noise.} When taking the product between the white noise and any stationary kernel, we recover another noise kernel. These redundant expressions are omitted from KITT's vocabulary. However, the result of the product between the noise kernel and the (non-stationary) linear kernel merits special attention. The linear kernel is defined as
\begin{equation}
    K_{\mathrm{LIN}}(\bm{x}, \bm{x'}) = \bm{\sigma}^2 (\bm{x} - \bm{c})(\bm{x'} - \bm{c}) \,  ,
\end{equation}
where $\bm{\sigma}^2$ denotes the vector of variances for the linear kernel, and $\bm{c}$ represents the shift parameter. Unlike the other primitive kernels, the linear kernel possesses independent variance terms for each input dimension.

The product between the linear kernel and the noise kernel generates a form of noise whose variance changes linearly with respect to the inputs. This presents an opportunity to model heteroscedastic noise, within the conventionally homoscedastic domain of GP regression models. When modelling real world tasks, this potentially offers a major advantage, since the noise variance often changes across the input space.  

Note that if we simply set $c=0$, as is often assumed when working with the linear kernel, then the linearly varying noise term is extremely limited. The noise variance could only ever increase as we move away from the origin. As with the cosine kernel, this reduces us to a small fraction $(2^{1-D})$ of the viable parameter space.  In order to lift this restriction, we introduce a `shift' vector in the linear kernel, such that the origin is free to move along each input dimension. This naturally leads us to ask what an appropriate prior for this shift vector would be. We consider it equally likely that the noise amplitude increases or decreases with $x$. To reflect this belief, and accounting for our normalised inputs, we seek a prior of the form $d\sigma^2/dx \sim \mathcal{N}(0, 0.1)$.  For large displacements, we note that the shift parameter can be approximately expressed as the ratio of two normally distributed variables: the gradient of the noise and the noise amplitude at the origin. This observation suggests a suitable prior on $c$ is given by the Cauchy distribution. We adopt $c \sim \mathrm{Cauchy}(0, 5)$ throughout.

\begin{figure}[t]
\centering
\includegraphics[width=\textwidth]{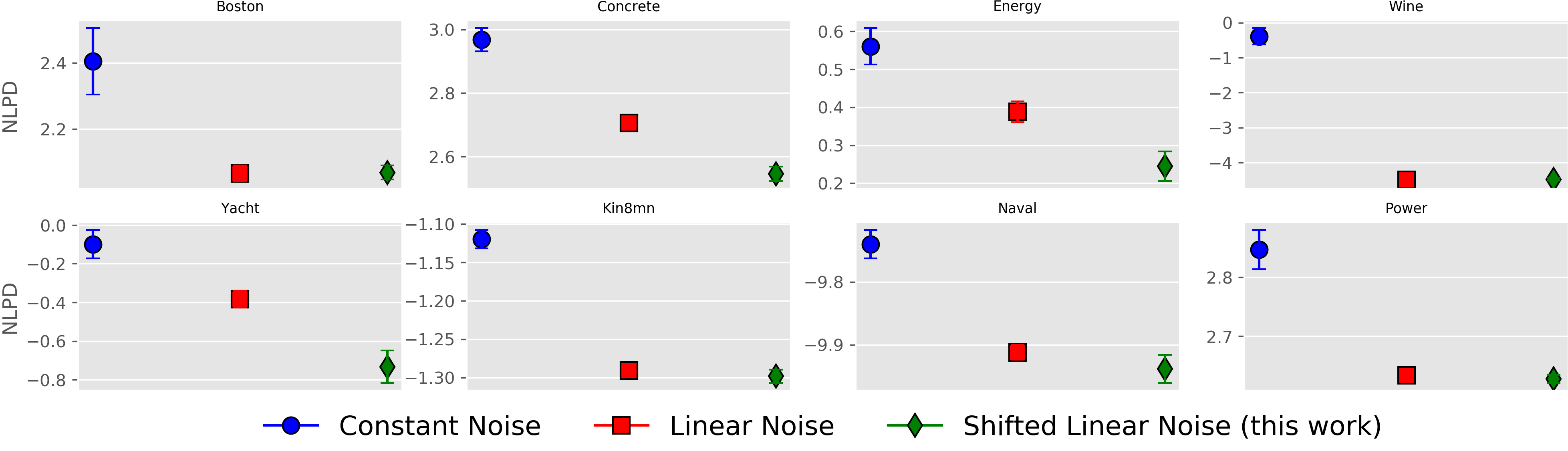}
\caption{Negative log likelihood values for three different noise models when used alongside the RBF kernel. All of the datasets clearly benefit from the modelling of heterescedastic noise, while three benefit from the additional freedom offered by the shift parameter.}
\label{fig:linear}
\end{figure}

\paragraph{Model Architecture.} 

The kernel's likelihood $p(\mathcal{D}|K)$ is invariant to permutations of the ordering of the datapoints, and to permutations of the ordering of the input dimensions. 
KITT's output of kernel recommendations should therefore exhibit these two important properties. \citet{zaheer2017deep} demonstrated that a function $f(X)$ which is invariant to permutations in the elements of $X$ can be expressed in the form $\rho \left(\sum_i \phi(x_i)) \right) $, where $\rho$ and $\phi$ are differentiable functions. This was exploited by \citet{lee2019set} in constructing the Set Transformer.
The scenario we encounter is slightly more complex in that there is a two-tiered hierarchy of invariances. We seek a function over the training set $\mathcal{D}$, which can be expressed as a collection of high dimensional data vectors $D_i$: $f(\left\{ \mathcal{D} \right\}) = \rho \left[\sum_i \phi(\mathcal{D}_i) \right]$.  Here the function $\phi(\mathcal{D}_i)$ must be invariant over the permutations of the different input dimensions. This can therefore be decomposed in a similar manner,
$\phi(D_i) =  \rho' \left[\sum_j \phi'(D_{ij}) \right]$.
Combining these two equations leaves a final expression of the form
\begin{equation}
    f(\left\{ \mathcal{D} \right\}) = \rho \left(\sum_i \rho' \left(\sum_j \phi'(\mathcal{D}_{ij}) \right) \right) \, , 
\end{equation} 
where $i$ and $j$ can be either dimensions or datapoints. This can be interpreted as: encode over $j$; pool over $j$; encode over $i$; pool over $i$; decode. This formalism sets the foundations for our choice of architecture, as shown in Figure \ref{fig:architecture}.  At a lower level, KITT is comprised of the following components whose acronyms we define here:

\begin{tabular}{ll}

$\bullet$ rFF: Row-wise feed-forward layer with ReLU & $\bullet$ MP: Mean pooling function \\
$\bullet$ SAB: Set Attention Block \cite{lee2019set} & $\bullet$ LayerNorm: Layer Normalisation \cite{ba2016layer} \\
$\bullet$ Multihead: Multi-Headed Attention Mechanism \cite{vaswani2017attention}.
\end{tabular}

\textbf{Encoder:} The encoder has the architecture of a transformer with self attention blocks. Our goal is to encode datasets $\mathcal{D}_{j} = \{\bm{x}_{i}, y_{i}\}_{i=1}^{N}$ of shape $N \times (D + 1)$ where $\bm{x}_{i} \in \mathbb{R}^{D}$ and $y_{i} \in \mathbb{R}$; we need to incorporate both invariance to ordering of the data points (row-wise shuffle) and equivariance to a re-ordering of the dimensions (column-wise shuffle). In order to achieve this, our encoder has two sub-components responsible for encoding along the sequence and dimension axes respectively:

\begin{enumerate}[labelindent=0pt, leftmargin=*]
\item \textbf{SEQ\_ENC:} $\mathbb{R}^{N \times (D + 1)} \rightarrow \mathbb{R}^{D \times E}$. A sequence encoding component which acts on input datasets and outputs dimension-level representations $\mathrm{G} \equiv \{\bm{g}_{d}\}_{d=1}^{D}, 
\bm{g}_{d} \in \mathbb{R}^{E}$ where $E$ is the embedding dimension. The sequence encoder forward pass is formulated as:
\begin{center}
SEQ\_ENC$(\mathcal{D})$ = MP(SAB$_{\times 6}(\text{rFF}(\mathcal{D})))$
\end{center}

Where mean pooling is applied over the sequence. Our implementation of the SAB component \cite{lee2019set} differs  slightly from that of \citet{lee2019set}, SAB$($Z$)$ = LayerNorm$($C + $Z)$, where we have defined $C = \mathrm{Dropout}(\mathrm{rFF}[\mathrm{Multihead}(Z,Z,Z)])$ \cite{vaswani2017attention}.

\item \textbf{DIM\_ENC:} $\mathbb{R}^{D \times E} \rightarrow \mathbb{R}^{D \times E}$ A dimension encoding component: DIM\_ENC which acts on dimension level encodings to generate final representations $\{\bm{h}_{d}\}_{d=1}^{D}$ of dimension $E$. The dimension encoder forward pass is formulated as:

\begin{center}
$\text{DIM\_ENC}(\text{G}) = \text{rFF}_{\times 2}(\text{SAB}_{\times 6}(\text{G}))$
\end{center}
\end{enumerate} 

The encoder forward pass entails passing each input data set to the sequence encoding component followed by the dimension encoding component. 

\begin{center}
$\text{ENCODER}(\mathcal{D}) = \text{DIM\_ENC} ( \text{SEQ\_ENC}(\mathcal{D}))$
\end{center}

\textbf{Proposition 1:} \textit{Let $\mathcal{S}_{N}$ denote the set of all permutations of the row-wise indices $\{1,2,\ldots,N \}$ and $\mathcal{D}_{\pi}$ denote an input dataset with the ordering of indices given by $\pi \in \mathcal{S}_{N}$ The sequence encoding component $\text{SEQ\_ENC}$ is invariant to a permutation of the indices within a dataset $\mathcal{D}$.}

\begin{center}
    $\text{SEQ\_ENC}(\mathcal{D}) = \text{SEQ\_ENC}(\mathcal{D}_{\pi})\ \forall\ \pi \in \mathcal{S}_{N}$
\end{center}

\textbf{Proposition 2:} \textit{Let $\mathcal{Q}_{D}$ denote the set of all permutations of the column-wise indices (dimensions) $\{1,2,\ldots,D\}$ and $\mathcal{D}_{\nu}$ denote an input dataset with the ordering of indices given by an arbitrary $\nu \in \mathcal{Q}_{D}$. The sequence encoder $\text{SEQ\_ENC}$ and dimension encoder $\text{DIM\_ENC}$ components are equivariant to a permutation of the dimensions within a dataset $\mathcal{D}$.}

\begin{center}
    $\text{ENCODER}(\mathcal{D}_{\nu}) = \nu ( \text{DIM\_ENC} ( \text{SEQ\_ENC} ( \mathcal{D} ) ) )\ \forall\ \nu \in \mathcal{Q}_D$
\end{center}
We include proofs for propositions 1 and 2 in the supplementary.

\textbf{Decoder:} KITT's decoder iteratively builds a caption from the encoded dataset representations, generated by the encoder described above, and a prompt which consists of the kernel expression thus far (see Fig. \ref{fig:architecture}). Our decoder closely resembles the one proposed in \citet{vaswani2017attention} except that we remove the positional encodings and adjust the number of layers.
The decoder uses self-attention blocks to first attend to the prompt and then to attend to the representations from the encoder using the processed dataset representations as query values. These two applications of attention are alternated in several layers until a new component kernel is proposed from a distribution generated from the final representations. We note that this component kernel proposition is invariant to the ordering of the dataset representations and thus to a shuffling of the dimensions of the original dataset. Hence, the model is fully invariant. The end-to-end process is depicted in Figure \ref{fig:architecture} and a detailed schematic of the decoder is included in the supplementary material.

\paragraph{Scalability.} 
Training of KITT is only performed once. Once training of the network is completed, all inference procedures require only a single forward pass through the network. 
As a result, instead of the $\mathcal{O}(N^3)$ cost commonly associated with explicit marginal likelihood evaluations, the cost of a forward pass through KITT is $\mathcal{O}(DN^2 + D^2)$. Due to effective parallelisation of the attention mechanism on GPUs, noted by \citet{vaswani2017attention}, and the modest size of the KITT network, we experience near constant wall clock time in practice.

\paragraph{Inference.} Given some new dataset $\mathcal{D}$, inference proceeds as below,  closely mimicking the procedure used to generate a caption for images. Note that the best caption is not necessarily constructed by choosing the best kernel at each step of the decoder.   
\begin{enumerate}
 
    \item Pass the data $\mathcal{D}$ into the encoder; pass the resulting encodings and an (initially empty) kernel expression $\mathcal{E}$ into the decoder.

    \item Retrieve the output probabilities, selecting the kernel $k$ with the highest probability and append the chosen kernel (or operator) to our full kernel expression $\mathcal{E}$. 

    \item Repeat steps 1 \& 2 until either the $< \!  \mathrm{STOP} \! >$ token is selected or the maximum caption length is reached.

    \item Repeat steps 1-3 several times but now select kernels stochastically, weighted by their probability, to construct a set of high-ranking kernel expressions.

    \item For each of the top three candidate kernels expressions, as ranked by their total probability assigned by the network, we optimise the associated hyperparameters with BFGS \citep{fletcher2013practical}.

    \item Combine the posterior distribution of the forecasts based on either the overall probability assigned by the network, or another proxy for the model evidence such as the Bayesian Information Criterion.
\end{enumerate}

In summary, we select a kernel based upon the output of the pretrained KITT network, before optimising the hyperparameters in a conventional manner.

\begin{figure}[t]
\centering
\includegraphics[width=\textwidth]{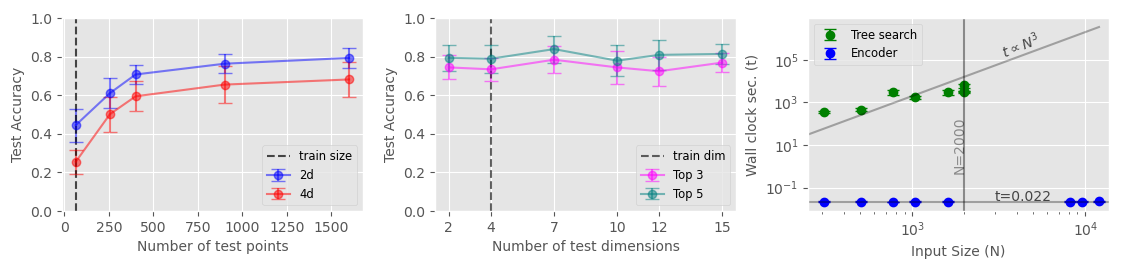} 
\caption[fontsize=small]{\emph{Left and centre}: Classification performance for random samples drawn from primitive kernels across a range of test sizes and dimensionality. The vertical dashed lines denote the conditions under which the network was trained. \emph{Right:} The time taken to predict a kernel for each of the UCI datasets. While KITT's overhead remains approximately constant, the tree search becomes impractical for larger inputs.} 
\label{fig:gtr}
\end{figure}

\section{Experimental Results}
\label{exp}
In this section we explore KITT's ability to predict kernels for synthetic data and standard regression benchmarks.  We present four baselines to assess the performance of KITT on regression data. AHGP \cite{liu2020task} is another deep network designed to assist GP inference, as outlined in \ref{bg}. 
The neural kernel network \cite{sun2018differentiable}, offers a differentiable form of kernel composition. We also include a greedy kernel search algorithm based upon the \emph{Automatic Statistician} procedure outlined in \citet{duvenaud2013structure}. These three algorithms span a wide range of computational overheads, with AHGP being the fastest and the kernel search being the slowest. As a more familiar reference point, we also include the RBF-ARD baseline, which uses the same priors described previously. 

\paragraph{Ground Truth Recovery.}
\label{gtr}
Identifying primitive structure is an important building block in being able produce sensible kernel recommendations for real-data in high dimensions. Capturing this structure is the aim of our encoder. In order to test the ability of the encoder to capture primitive structure, we form a classification transformer by taking the KITT encoder and appending a dense layer followed by a softmax activation. The resulting model is trained on datasets of fixed size and dimensionality to predict kernels for synthetic datasets drawn from GPs with known kernels.
We demonstrate test performance in terms of accuracy for varying test input sizes and dimensions. We draw 300 random samples from a selection of primitive kernels, for varying combinations of test size and dimensionality.
The results shown in Figure \ref{fig:gtr} demonstrate that the classifier is able to reliably generalise its structure detection capabilities to higher dimensional tasks, which were unseen during training of the network.  As is expected, a moderately sized dataset of at least $\sim 200$ points is needed to achieve a reasonable level of prediction accuracy, and this continues to improve with increasingly large datasets.  
\begin{figure}[t]
\centering
\includegraphics[width=\textwidth]{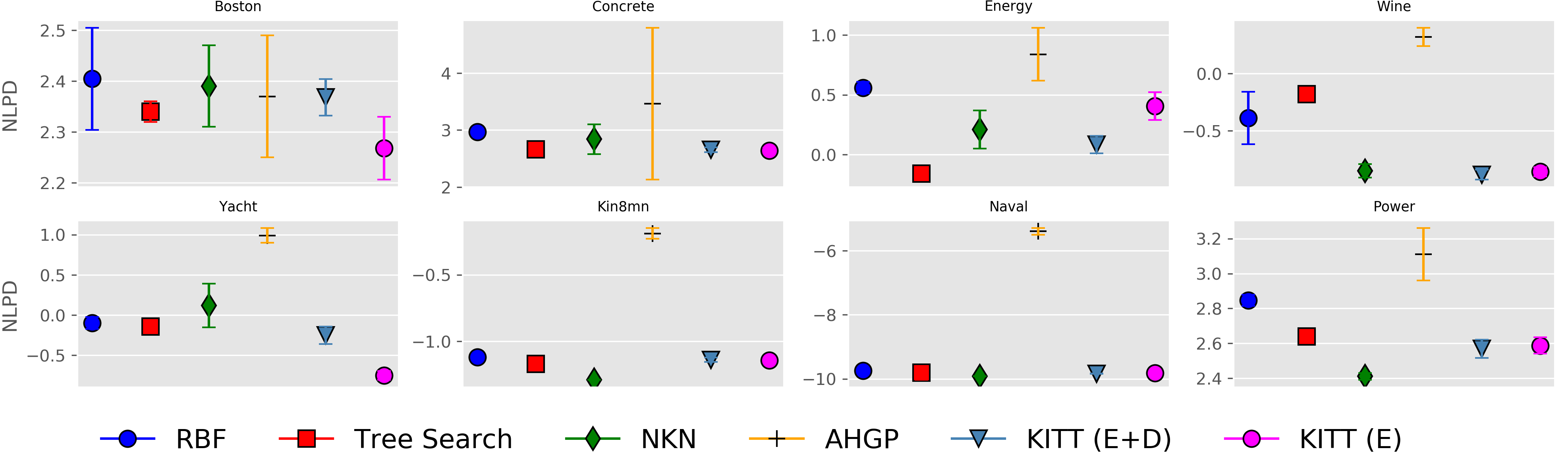} 
\caption{Negative predictive log likelihood values on UCI regression tasks for a variety of kernel selection methods. KITT remains competitive with the most computationally intensive approach, the tree search, while offering the advantage of being several orders of magnitude faster. The `E' denotes sole use of the encoder followed by a classification layer to select kernels, while `E+D' generates captions with the decoder. }
\label{fig:nlpd_uci}
\end{figure}

\paragraph{UCI Regression.}

We evaluate KITT on eight real-world UCI regression tasks\footnote{A recurring misconception in the literature is that the predictive errors on the Naval dataset are so small that they may be neglected, and these are sometimes listed as ``0.00''. We stress that they are small only because of the small variance of the raw data. This should have no bearing on its significance alongside the other datasets, and the RMSE should not be rounded down to zero.}, spanning  a range of input sizes and input dimensions (from 4 to 14). We adopt the same benchmarking methodology as \citet{liu2020task}, which includes a 90/10 train/test split, and subsampling 2,000 datapoints for those cases where the dataset exceeds this number. For each dataset, we predict a kernel caption with a maximum expression length of three terms. The caption is either constructed sequentially by the decoder, or in the case of the classifier, by summing the three highest scoring kernels.

The resulting NLPD values from KITT, and three other approaches to kernel design, are shown in Figure \ref{fig:nlpd_uci}. Uncertainties are estimated from repeated experiments with ten different splits.  KITT consistently outperforms the other transformer-based model, AHGP, and is competitive with the far slower tree search method (see Figure \ref{fig:gtr}). We also perform an ablation study, details of which can be found in the supplementary material, demonstrating that KITT outperforms a random selection from its vocabulary. 

We note that the AHGP performance is weaker than that of the RBF. This is significant because the RBF is a subset of the Spectral Mixture Product model, equivalent to a single component set to zero mean frequency. This suggests that the AHGP network perhaps focused on learning how to identify the lengthscale of the primary component of the multi-component Spectral Mixture Product. 

\begin{figure}[t]
\centering
\includegraphics[scale=0.35]{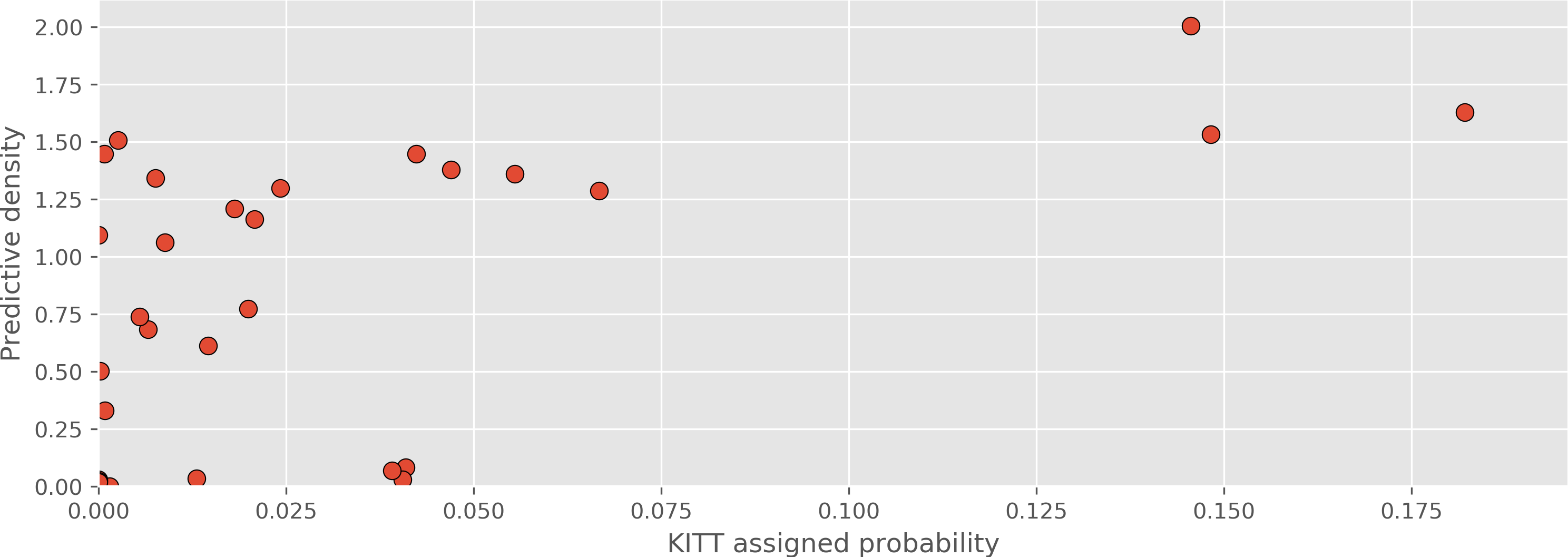}
\caption{A comparison of KITT's kernel predictions against their test performance, on the Yacht dataset. Each dot represents one of the 34 kernels in KITT's vocabulary. KITT successfully identifies the three top performing kernels, and assigns low probability to the 31 alternative options. }
\label{fig:yacht}
\end{figure}

For a deeper understanding of KITT's performance, Figure \ref{fig:yacht} compares the network's output against realised test performance on the Yacht dataset, across all 34 kernel classes. The three kernels KITT assigned high probability to, namely $\text{Linear}\times\text{RBF}$, $\text{Linear}\times\text{Matern32}$ and  $\text{Linear}\times\text{Matern52}$, correspond to the three strongest test performances. 

\paragraph{Computational overhead.}
One of the most compelling features of KITT is the speed at which inference can be performed. Identifying a suitable kernel for a previously unseen dataset only entails a single forward pass through the encoder, and a small number through the decoder, each of which requires around two hundredths of a second. Furthermore, as illustrated in the right hand panel of Figure \ref{fig:gtr}, the time-cost of prediction is robust to increasing data-set sizes and dimensionality.

The KITT network was trained on a Tesla V100 GPU for approximately eight hours, with Adam \citep{kingma2014adam}. This procedure occurred only once, and does not need to be repeated when performing inference. To generate a kernel prediction requires a small fraction of a second, and is largely insensitive to the size of the input data, as seen in the right hand panel of Figure \ref{fig:gtr}. Once a kernel has been recommended, training typically required a further ten seconds.  It is possible this step could also be greatly accelerated in future, if KITT were used in tandem with a hyperparameter optimisation network, similar to AHGP \cite{liu2020task}.

\section{Related work}

In this section we review approaches that either directly, through kernel construction, or indirectly target the issue of model selection in GPs.

\textbf{Amortised Hyperparmeter Learning (AHGP):} \citet{liu2020task} also use self-attention based transformers, but with the goal of amortising hyperparameter learning. They train on input-output regression based datasets to estimate the final set of GP hyperparameters that would otherwise be learnt as a result of maximizing the marginal likelihood. In order to circumvent kerrnel selection they choose the flexible spectral mixture (SM) kernel with a fixed number of components per dimension, yielding a kernel with product structure over dimensions. The SM kernel arises from modelling the spectral density (Fourier dual) of a kernel function as a Gaussian mixture. Our work differs from AHGP as we focus on kernel design rather than optimising hyperparameter values. 

\textbf{Kernel Engineering:} 
There are several examples in the GP literature of kernels being handcrafted to model one- or two-dimensional data \citep{duvenaud2014automatic, duvenaud2013structure, duvenaud2011additive, chapter4}. While this may be feasible in low-dimensional data with the aid of visual inspection, it is much less straightforward in a high-dimensional settings. Automated kernel engineering approaches search over a finite space of kernel structures which are progressively built by adding and multiplying a small number of base kernels. The focus is on devising an effective search algorithm over discrete structures where the end result is a composite kernel built from simpler known base kernels. This is largely the idea behind the \textit{Automatic Statistician} project \citep{kim2018a} where a greedy search procedure searches over all possible operators and sub-expressions to select the highest scoring combination. Our work similarly operates on a universe of compositional kernel structures but with a distinctly different model where we regress kernel labels on data sets with end-to-end gradient based training yielding a fast and scalable method.

\textbf{Deep Kernels:} There are other methods that bring to bear both the benefits of deep architecture and the analytical flexibility of kernel methods
for the problem of representation learning 
\citep{calandra2016manifold, hinton2008using, wilson2016deep}. The methods work by transforming the inputs to a GP with a neural network (NN) and jointly learning the parameters of the NN and the GP. The contention is that a simple base kernel (like a squared exponential (SE) kernel) works better when applied to the representations learnt by the NN than when applied to the raw input. These works try to side-step the problem of learning a sophisticated kernel apt for the data by focusing instead on learning a transformation of inputs. However, these methods can suffer from overfitting due to the joint training of millions of parameters of the NN in conjunction with the GP hyperparameters \citep{ober2021promises}.

\textbf{Novel Kernels:} Other noted work includes the spectral mixture kernel which reparameterizes the kernel in terms of its spectral density (see \emph{Bochner's Theorem} \citep{bochner1959lectures}) and derives closed form kernels which can be used as drop-in replacements for any stationary kernel function \citep{simpson2021minecraft, wilson2014covariance}.

\section{Discussion}

This work proposes a novel approach to addressing the kernel selection problem in GPs. By leveraging the potential for unlimited training data, we train a transformer-based model to identify the likelihood of a sample given a kernel class. Despite being trained solely on synthetic data, KITT is capable of selecting suitable  kernels for previously unseen, real-world datasets. While we focus our efforts on the case of one-dimensional outputs, similar models could be developed for multi-output regression, classification, latent variable modelling and time-series prediction tasks. A major advantage of a pre-trained model for kernel structure detection is the speed of inference. By being able to recommend a kernel in a fraction of a second, KITT is dramatically faster than competing methods such as greedy search algorithms or differentiable kernel networks. Furthermore, it offers superior scalability.
Empirically, we found that KITT is capable of pattern discovery across a broad range of input dimensions and dataset sizes. It was found to predict competitive kernels for high-dimensional real valued regression tasks. The ground truth experiments demonstrate its generalisation ability where it is able to identify structure in high-dimensional datasets. This work presents a powerful hybrid approach where kernel selection is informed by representation learning, by inferring a range of kernels compatible with the data. This achieves two aims of expressivity and ensemble uncertainty while spurring new possibilities for informed model selection in Gaussian processes. Given the high degree of complementarity with AHGP, which offers near-instantaneous optimisation of hyperparamters, there appears promising prospects for transformers to enhance the development, flexibility and scalability of Gaussian Process models.

\begin{ack}
The authors would like to thank the anonymous reviewers for their helpful feedback. 
VL acknowledges funding from the Alan Turing Institute and Qualcomm Innovation Fellowship (Europe). 
\end{ack}

\bibliographystyle{plainnat}
\bibliography{kitt}

\end{document}


\maketitle
\appendix

\section{Background: Self-Attention}
\label{sa}

Since the attention mechanism is rarely used within the GP literature, we provide a brief review of the topic in this section.  Below we follow the description of attention as given by \citet{vaswani2017attention}, including extensions to self-attention and multi-head self-attention. 

The dot-product attention mechanism \citep{vaswani2017attention} takes as input a set of queries, keys and values. The queries and keys have dimension $D_z$ and the values have dimension $D_v$ which may differ from $D_z$. The operation of dot-product attention then generates weights from the queries and keys which are used to produce a linear mapping of the input values.
\begin{equation}
    Attention(Q,K,V) = \text{softmax}\left(\frac{QK^\top}{\sqrt{D_z}}\right)V \, ,
\end{equation}
where the $Q$, $K$ and $V$ matrices denote the row-wise collection of queries, keys and values respectively. The softmax operation is applied row-wise with scaling to avoid the inputs exploding which hampers training. Intuitively, attention enables the input values to be processed to yield representations which account for the context of all other values in the sequence.

Dot-product self-attention acts on a single sequence of inputs, using it to generate queries, keys and values for the attention mechanism described above. Queries, Keys and Values are generated by right multiplication with learned weight matrices, $W_Q$, $W_K$ and $W_V$ respectively. Self-attention is therefore a mathematical operation which takes as input, a set of vectors of length $D$ collected row-wise in $Z$, $Z \equiv \{z_{i}\}_{i=1}^{N}, z_{i} \in \mathbb{R}^{D}$, and computes a weighted sum of vectors for each index $i$, 
\begin{equation}
    y_{i} = \sum_{j}w_{ij}z_{j} \, .
\end{equation}
The weights, which collectively form a matrix $W$, are a function of the input vectors in $Z$, 

\begin{equation}
W = \textrm{softmax}\left(\hat{W}_i/\sqrt{D}\right)\quad\text{where}\quad \hat{W} = ZW_QW_K^\top Z^\top \, .
\end{equation}

When this operation is repeated $H$ times with individual sets of matrices $\{W_{Q}^{h}, W_{K}^{h}, W_{V}^{h}\}_{h=1}^{H}$ the resulting operation is called \textit{Multi-Head Self Attention} (MHSA). Let $a_{h}$ denote the attention encoded output of a single head, the output of multi-head attention is then computed as,  

\begin{equation}
\textrm{MHSA}(Z) \equiv \mathrm{Multihead}(Z, Z, Z) = \underbrace{\textrm{concat}(a_{1}, ...a_{H})}_{N \times HD}W_{0} \in \mathbb{R}^{D} \, ,
\end{equation}
\begin{equation}
\textrm{where } a_{h} = Attention_{h}(ZW_{Q}^{h}, ZW_{K}^{h}, ZW_{V}^{h}) \, .
\end{equation}
Here $W_{0} \in \mathbb{R}^{HD \times D}$ is the final trainable projection matrix which brings the outputs of the individual heads back to the original dimension $D$.


\section{Proofs}

\subsection{On encoding sets of datasets}

Our transformer architecture is trained on input-output datasets where the output is drawn from GP priors with a known kernel. Each labeled example is thus a self-contained dataset.
This introduces some challenges in ensuring that each dataset's encoding is not sensitive to the ordering of the data points within each dataset
. A training instance $\mathcal{D} = \{\{\bm{x}_{i,j}\}_{j=1}^{D},y_{i} \}_{i=1}^{N} \in \mathbb{R}^{N\times D + 1}$ denotes a rank-2 input tensor which carries the interpretation of a \textbf{set} of $N$ vectors in $\mathbb{R}^{D + 1}$. For instance, $[x_{i,1}, \ldots, x_{i,D}, y_{i}]$ is the $i^{th}$ vector. Hence, we wish to preserve invariance of the encoded representation with respect to a permutation of the rows within each dataset. 
\begin{equation}
    \textrm{ENCODER}(\mathcal{D}_{\pi}) = \textrm{ENCODER}(\mathcal{D})
\end{equation}
where $\mathcal{D}_{\pi} = \{\{\bm{x}_{\pi(i),j}\}_{j=1}^{D},y_{i} \}_{i=1}^{N}, \textrm{ and } \pi: \{1, \ldots, N\} \rightarrow \{1, \ldots, N\}$ is a bijective permutation function on the row indices. This is addressed by Proposition 1.

Further, we want to ensure that the encodings of each dataset are equivariant to the ordering of the input dimensions (columns) in each dataset. If $\nu: \{1, \ldots, D\} \rightarrow \{1, \ldots, D\}$ is a bijective permutation function on the column indices then, 
\begin{equation}
    \textrm{ENCODER}(\mathcal{D}_{\nu}) = \nu(\textrm{ENCODER}(\mathcal{D}))
\end{equation}
where $\mathcal{D}_{\nu} = \{\{\bm{x}_{i,\nu(j)}\}_{j=1}^{D},y_{i} \}_{i=1}^{N}$. Note that the position of dimension $(D + 1)$ denoting the output column in each training instance is always preserved. We address this permutation equivariance in Proposition 2.

\textbf{Lemma 1:} The self-attention mechanism (for each head) defined in \ref{sa} is permutation equivariant. 
$$ \textrm{softmax}\left(\dfrac{Z_{\pi}W_{Q}W_{{K}}^{T}Z_{\pi}^{T}}{\sqrt{D}}\right)Z_{\pi}W_{V} = \pi\left(\textrm{softmax}\left(\dfrac{ZW_{Q}W_{{K}}^{T}Z^{T}}{\sqrt{D}}\right)ZW_{V}\right) \forall \ \pi \in S_N$$

where $Z_{\pi} = \pi(Z)$ denotes a permutation of the rows in $Z, \pi \in S_{N}$ where $S_{N}$ denotes the set of all permutations of the row indices $\{1,\ldots,N\}$.
\begin{proof}
First, we note that the softmax and scaling are element-wise operations and don't interfere with the ordering of rows; in order to 
prove the permutation equivariance we just need to focus on the matrix multiplication operations.

To prove:
\begin{equation}
    Z_{\pi}JZ_{\pi}^{T}Z_{\pi}W_{V} = \pi(ZJZ^{T}ZW_{V})
    \label{cond}
\end{equation}

where we have pre-multiplied $W_{Q}W_{K}^{T} = J$ (by the associativity of matrix multiplication).

Without loss of generality assume,

\begin{equation}
Z = \begin{bmatrix} 
a & b \\
c & d \\
\end{bmatrix} \hspace{4mm} Z_{\pi} = \begin{bmatrix} 
c & d \\
a & b \\
\end{bmatrix} \hspace{4mm} Z_{\pi}^{T} = \begin{bmatrix} 
c & a \\
d & b \\
\end{bmatrix}\hspace{4mm}  J =  \begin{bmatrix} 
j_{1} & j_{2} \\
j_{3} & j_{4} \\
\end{bmatrix} \hspace{4mm} W_{V} = \begin{bmatrix}
v_{1} & v_{2} \\
v_{3} & v_{4} \\
\end{bmatrix}
\end{equation}

First, we note that the operation $Z^{T}Z$ is permutation invariant, $Z_{\pi}^{T}Z_{\pi} = Z^{T}{Z}$
$$ Z_{\pi}^{T}Z_{\pi} = \begin{bmatrix} 
c & a \\
d & b \\
\end{bmatrix} \begin{bmatrix} 
c & d \\
a & b \\
\end{bmatrix} = \begin{bmatrix} 
c^{2} + a^{2} & cd + ab \\
cd + ab & b^{2} + d^{2} \end{bmatrix} = \begin{bmatrix} 
a & c \\
b & d \\
\end{bmatrix} \begin{bmatrix} 
a & b \\
c & d \\
\end{bmatrix} = Z^{T}Z      \\
$$
Hence, the LHS term in \eqref{cond} becomes, 
\begin{equation}
    Z_{\pi}JZ_{\pi}^{T}Z_{\pi}W_{V} = Z_{\pi}\underbrace{JZ^{T}ZW_{V}}_{H} = Z_{\pi}H = \pi(ZH) 
\end{equation}
where the final equality can be shown to be true by assuming without loss of generality, $$H = \begin{bmatrix} h_{1} & h_{2} \\
h_{3} & h_{4} \\
\end{bmatrix},$$
Hence,
\begin{align*}
    Z_{\pi}H = \begin{bmatrix} 
c & d \\
a & b \\
\end{bmatrix}\begin{bmatrix} h_{1} & h_{2} \\
h_{3} & h_{4} \\
\end{bmatrix} &= \begin{bmatrix}
ch_{1} + dh_{3} & ch_{2} + dh_{4} \\
ah_{1} + bh_{3} & ah_{2} + bh_{4} \\
\end{bmatrix} \\
&=\pi\left(  \begin{bmatrix}
ah_{1} + bh_{3} & ah_{2} + bh_{4} \\
ch_{1} + dh_{3} & ch_{2} + dh_{4} \\
\end{bmatrix}     \right)\\
&= \pi\left(  \begin{bmatrix}
a & b \\
c &  d \\
\end{bmatrix}   \begin{bmatrix} h_{1} & h_{2} \\
h_{3} & h_{4} \\
\end{bmatrix}  \right)\\ &= \pi(ZH)
\end{align*}
\end{proof}

Note that this proof shows equivariance of self-attention in a rank 2 input tensor case $(Z \in \mathbb{R}^{D \times D})$ however the proof can be generalised to higher rank inputs. For example, if $Z \in \mathbb{R}^{D\times E \times J}$ is a rank 3 input tensor and we permute the indices on the $E$ dimension $Z_{D\times E \times J} \rightarrow Z_{D \times \pi(E) \times J} = Z_{\pi^{E}}$, then, the output of self-attention is permuted on the same dimension. 

\begin{equation}
    \textrm{MHSA}(Z_{\pi^{E}}) = \pi^{E}(\textrm{MHSA}(Z))
\end{equation}

In the full encoder architecture, the attention mechanism is applied internally as part of a sequence of blocks called set attention blocks $\text{SAB}$\cite{lee2019set}\footnote{We note that our implementation of the set attention block differs slightly in its use of dropout and residual connections from that of \citet{lee2019set}.}. The block operation leaves the architecture permutation equivariant.

\textbf{Lemma 1.1:} \textit{The set attention block} SAB$(Z)$ \textit{is permutation equivariant.}
\label{mab}
\begin{center}
   \textrm{SAB}$(Z)$ := LayerNorm$(C + Z)$
\end{center}
where C denotes a context vector computed with the Attention Mechanism \ref{sa}.

\begin{center}
    C = rFF$($MHSA$(Z))$
\end{center}

\begin{proof}
We know that the $\textrm{MHSA}(\cdot)$ operation is permutation equivariant and LayerNorm is an independent element-wise operation with no parameters. It remains to verify that  the feed-forward operation leaves the outputs permutation equivariant. 

Without loss of generality assume that the outputs of MHSA$(Z)$ are given by an $N{\text -}D$ tensor $A \in \mathbb{R}^{D \times D ...\times D}$. The feed-forward layer with $k$ hidden units applies a matrix $W_{k}$ of dimension $D\times k$ along the last axis of the inputs yielding a tensor output $\textrm{rFF}(A) = AW_{k}$ of shape $D \times  \ldots D \times k$. Essentially, each sub-tensor of shape $(1 \times \ldots \times D)$ (row of size $D$) is multiplied by the weight matrix independently and identically to yield the output sub-tensors of shape $(1 \times \ldots 1 \times k)$ (row of size $k$). Since this operation is applied row-wise it is permutation equivariant to the order of the order of the sub-tensors in $A \Rightarrow \textrm{rFF}(A_{\pi}) = \pi(\textrm{rFF}(A))$.
\end{proof}

\textbf{Proposition 1:} \textit{Let $\mathcal{S}_{N}$ denote the set of all permutations of the row-wise indices $\{1,2,\ldots,N \}$ and $\mathcal{D}_{\pi}$ denote an input dataset with the ordering of indices given by $\pi \in \mathcal{S}_{N}$ The sequence encoding component} SEQ\_ENC \textit{ is invariant to a permutation of the indices within a dataset $\mathcal{D}$.}

\begin{center}
    SEQ\_ENC$(\mathcal{D})$ = SEQ\_ENC$(\mathcal{D}_{\pi})\ \forall\ \pi$ in $\mathcal{S}_{N}$
\end{center}

\begin{proof}

The sequence encoder forward pass is formulated as:
\begin{center}
SEQ\_ENC$(\mathcal{D})$ = MP$($SAB$_{\times 6}($rFF$($R$(\mathcal{D}))))$
\end{center}

where $R$ is a reshape operation which takes a rank-2 tensor input of size $(N \times D + 1)$ and outputs a rank-3 tensor of shape $(D\times N \times 2)$. This reshaped tensor is formed by stacking row-wise of the $N \times 2$ sub-tensors corresponding to each dimension yielding $\{\{(x_{i,j},y_i)\}_{i=1}^N\}_{j=1}^D$. Let $\mathcal{D}_{\pi}$ denote a training instance (a dataset) of shape
$(N\times D + 1)$ where $\pi$
denotes a permutation of the rows. The output of the reshape operation $R$ is a rank-3 input tensor of shape $D \times \pi(N) \times 2$ where the order of data points has been permuted for each dimension. rFF is permutation equivariant as it acts on rows of the input dataset, hence $\text{rFF}(\text{R}(\mathcal{D}_{\pi})) = \pi(\text{rFF}(\text{R}(\mathcal{D})))$.
The SAB block \citep{lee2019set} is permutation equivariant from Lemma 1.1. We know that permutation equivariant layers stacked together are permutation equivariant \citep{zaheer2017deep}. Hence, a composition of SAB layers $(\text{SAB}_{\times6})$ with rFF is permutation equivariant. 

Further, the mean-pooling operation MP applied across the sequence $(N)$ in each dimension is permutation invariant by definition, hence,
\begin{align}
\text{SEQ\_ENC}(\mathcal{D}_{\pi}) &= \text{MP}(\text{SAB}_{\times 6}(\text{rFF}(\text{R}( \mathcal{D}_{\pi} )))) \\
&= \text{MP}(\pi(\text{SAB}_{\times 6}(\text{rFF}(\text{R}( \mathcal{D} )))) \\
&= \text{MP}(\text{SAB}_{\times 6}(\text{rFF}(\text{R}( \mathcal{D} )))) \\
&= \text{SEQ\_ENC}(\mathcal{D})
\end{align}
\end{proof}

\textbf{Proposition 2:} \textit{Let $\mathcal{Q}_{D}$ denote the set of all permutations of the column-wise indices (dimensions) $\{1,2,\ldots,D\}$ and $\mathcal{D}_{\nu}$ denote an input dataset with the ordering of indices given by $\nu \in \mathcal{Q}_{D}$. The sequence encoder} SEQ\_ENC \textit{and dimension encoder} DIM\_ENC \textit{components are equivariant to a permutation of the dimensions within a dataset $\mathcal{D}$.}

\begin{equation*}
    \text{ENCODER}(\mathcal{D}_{\nu}) = \nu (\text{DIM\_ENC}(\text{SEQ\_ENC}( \mathcal{D} )))
    \ \forall\ \nu \in \mathcal{Q}_{D}
\end{equation*}

\begin{proof}
First, we tackle the sequence encoder SEQ\_ENC. Let $\mathcal{D}_{\nu}$ denote a training instance of shape
$(N \times D + 1)$ where $\nu$
denotes a permutation of the input columns\footnote{The output column is always on the last axis}. The output of the reshape operation is a 3d input tensor of shape $(D \times N \times 2)$. Since the feed-forward layer rFF applies to each $N \times 2$ sub-tensor independently and identically it renders the output permutation equivariant, $\text{rFF}(\text{R}(\mathcal{D}_{\nu})) = \nu (\text{rFF}(\text{R}(\mathcal{D})))$.

\begin{figure*}[t]
\centering
\includegraphics[width=0.9\textwidth]{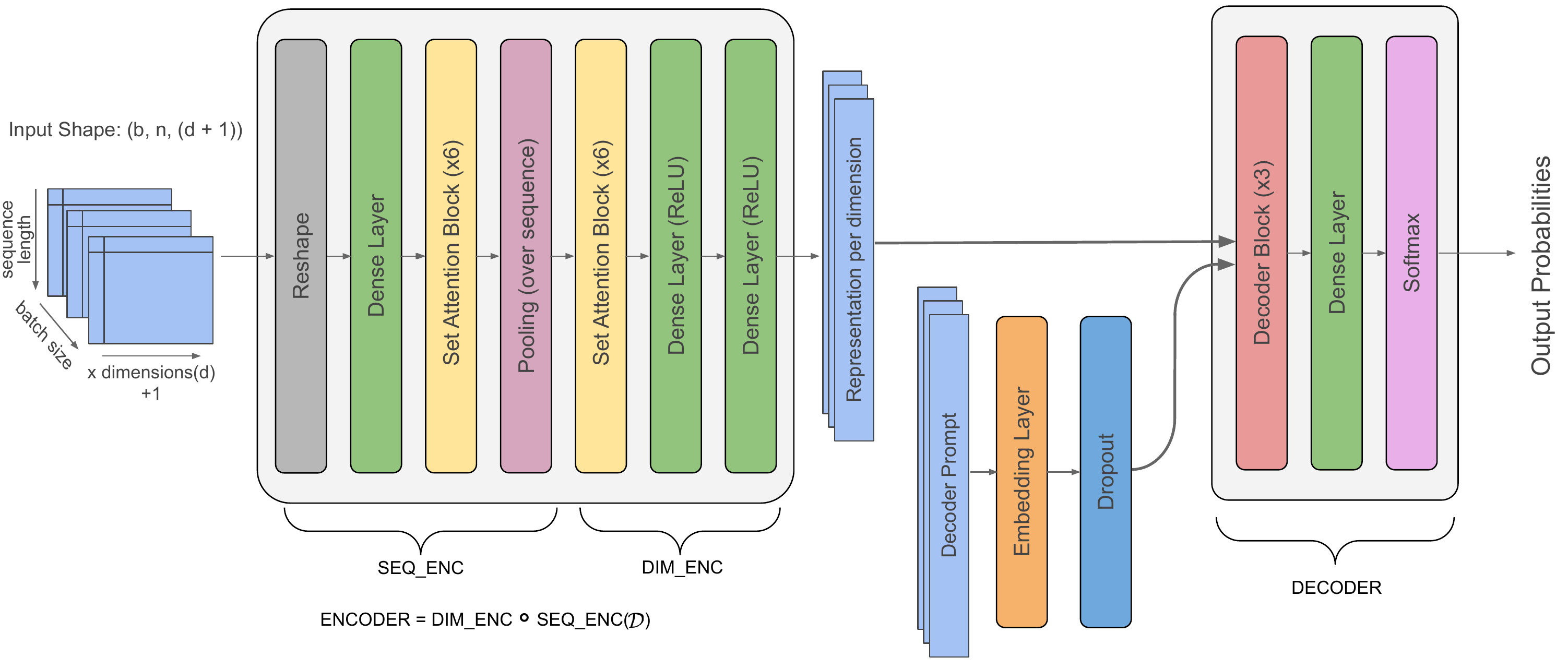}
\caption{KITT architecture adapted from image captioning network \citet{xu2015show}.}
\label{fig:detailed-architecture}
\end{figure*}

The SAB$_{\times 6}$ block is permutation equivariant from Lemma 1.1. The mean-pooling operation collapses the dimension of size $N$ (i.e. it is applied across the sequence of data points in each dimension) yielding an output which is permutation equivariant to the order of dimensions.

Hence, SEQ\_ENC$(\mathcal{D}_{\nu}) = \nu(\text{SEQ\_ENC}(\mathcal{D}))$.

The dimension encoder DIM\_ENC is implemented as a stack of multi-head attention blocks SAB$_{\times 6}$ which are permutation equivariant (as shown), hence the full encoder is a composition of permutation equivariant transformations (w.r.t. dimensions),
\begin{center}
    DIM\_ENC$($SEQ\_ENC$(\mathcal{D}_{\nu} )) = \nu ($DIM\_ENC$($SEQ\_ENC$( \mathcal{D} )))$
\end{center}
\vspace{-5mm}
\end{proof}
The encodings produced by our encoder are therefore invariant to the ordering of data points and equivariant with respect to permutations of dimension. When coupled with the fact that we do not add any location encoding before passing the encodings to our decoder, this renders the kernels proposed by the full KITT model fully invariant to permutations of both the dimensions and indices in the input dataset. This follows from the decoder treating all of the provided encodings equally, attending to them according to the values of the queries generated by the prompt provided and the keys calculated from the encodings themselves. More information on the architecture of the decoder is provided in the next section.

\section{KITT Architecture}

In this section, we expand on the discussion of KITT's architecture in the main body of the paper with additional detail and diagrams.

Figure \ref{fig:detailed-architecture} shows the full end-to-end KITT architecture which maps from a dataset to a distribution over kernels. Both the encoder and the decoder utilise sets of 6 attention blocks in conjunction with typical densely connected layers. Where no activation function is specified the dense layer simply represents multiplication by a matrix of learned weights. These dense layers operate only on the final dimension of the input tensors and are therefore referred to as row-wise feed forward layers (rFF).
As shown in Figure \ref{fig:architecture}, the operation of the decoder is recurrent in so far as it is run several times with the same encodings, updating the prompt with the proposed kernel from the last step at each iteration.
We only run the decoder multiple times, as it is only the prompt that changes each time; once calculated the encoded dataset representations are cached and reused in each pass of the decoder. Note that during training there is no need for repeated forward passes through the decoder as the full caption is known. We are therefore able to train the decoder by passing copies of the caption with differing masking through the decoder to effectively parallelise decoder prediction at each point of caption generation. The mask acts to prevent looking ahead to what is to be predicted.

\begin{figure}[t]
\centering
\includegraphics[scale=0.3]{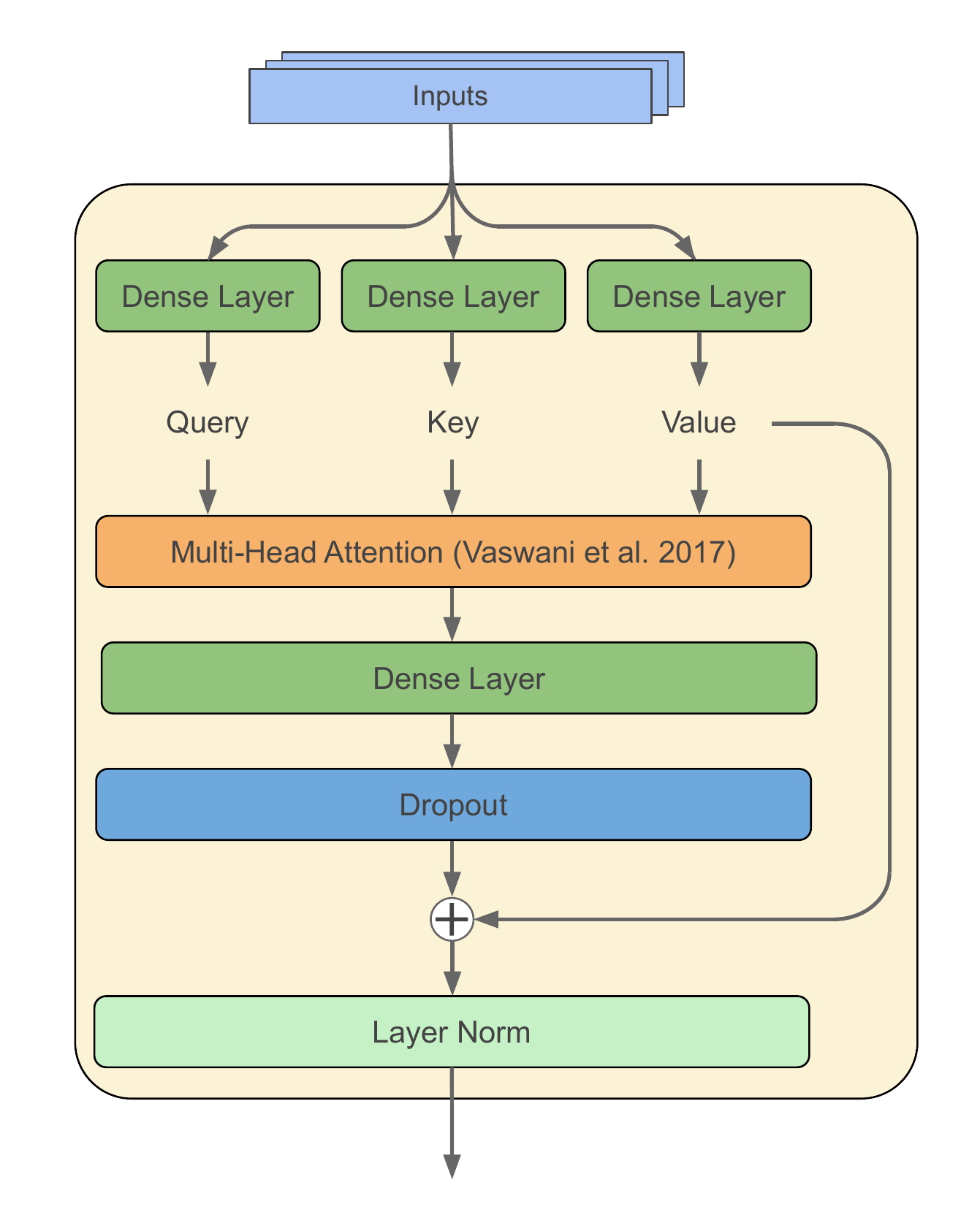}
\includegraphics[scale=0.27, angle=90]{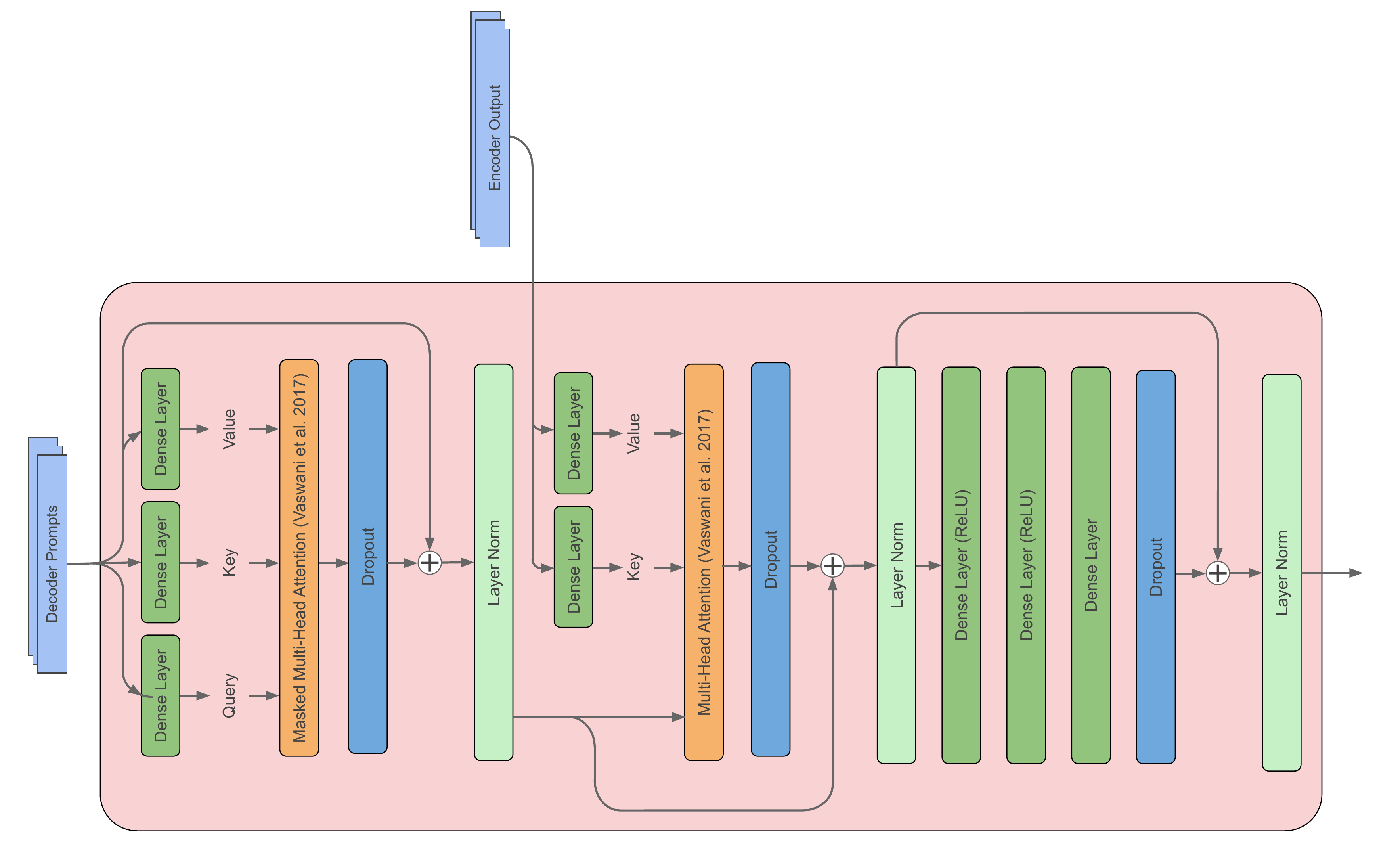}
\caption{\textit{Left:} Set Attention Block. \textit{Right:} Decoder Block}
\label{fig:mhsa-block}
\end{figure}

The internal operations of the Set Attention Blocks and the Decoder Block which form the core of the encoder and decoder respectively are depicted in Figure \ref{fig:mhsa-block}.

Our Set Attention Blocks differ slightly from those introduced by \citet{lee2019set}. Following \citet{liu2020task}, we utilise dropout \cite{gal2016dropout} with a rate of $0.1$, and favour a single residual connection and layernorm \cite{ba2016layer} operation as compared to the two instances of layer norm with residual connection used by \citet{lee2019set}.

A key difference between our encoder and the original transformer \cite{vaswani2017attention} is the absence of an explicit positional encoding added to the inputs. In our setting of processing full datasets, this is not required as the relevant positional information is naturally contained within the inputs. The position of a datapoint is defined by its $\boldsymbol{x}$ values which form an integral part of the input and are therefore processed directly by the transformer without any need for external processing. This denotes an important generalisation of the Transformer beyond it's original application in natural language processing to our setting of detecting patterns in any numerical dataset.

Our decoder blocks are implemented as originally proposed by \citet{vaswani2017attention} when introducing the Transformer architecture. However, unlike \citeauthor{vaswani2017attention}, we omit positional encodings from the input prompt. Positional encodings are not required when processing a kernel caption due to the transitivity of the addition of kernels ($k_a + k_b = k_b + k_a)$. The order of previously predicted kernels does therefore not influence the prediction of the next kernel. It is only the set of previous predictions which is important. In order to enable the ordering of kernels within a sum to convey some degree of information, we adopt a formalism where kernels are stated in decreasing order of their variance.

\section{Further Experimental Results}

Here we present further details of our experimental results. The NLPD values illustrated in Figure 4 in the main text, corresponding to the UCI regression tasks, are shown in Table \ref{tab:nlpd}. The RMSE values for the same experiments are given in Table \ref{tab:rmse}. 
Finally, Table \ref{tab:strats} shows that performing model averaging across the top three kernels offers some advantage over simply selecting the top one, while both approaches comfortably outperform a random kernel selection.  

In all cases, quoted uncertainties are estimated from repeating ten different splits.

\begin{table*}[t]
\resizebox{\columnwidth}{!}{%
\begin{tabular}{l|l|c|c}
Approach & Kernel & Hyperparameters  & Reference \\
\hline
Tree Search (TS) & Flexible & ML-II  & \citet{duvenaud2014automatic} \\
Neural Kernel Network (NKN) & Flexible & ML-II &  \citet{sun2018differentiable} \\
Amortised Hyper. Inference (AHGP) & Fixed (SM) & Amortised & \citet{liu2020task} \\ 
KITT (This work) & Flexible & ML-II & -- \\ 
\end{tabular}}
\caption{A summary of the kernel selection and optimisation methods under consideration.}
\label{tab:comp}
\end{table*}

\begin{table*}[t]
\centering

\begin{tabular}{l|c|c|c|c|c|c}
Dataset  & RBF & TS & NKN & AHGP & KITT (E + D) & KITT (E) \\
\hline
Boston & 2.40 (0.05) & 2.08 (0.02) & 2.39 (0.08) &  2.37 (0.12)  & 2.37(0.04) & 2.27 (0.06)\\
Concrete   & 2.96 (0.03) & 2.46 (0.03) & 2.84 (0.26) &  3.46 (1.33) & 2.65 (0.04) & 2.64 (0.04)\\
Energy  & 0.56 (0.04) & -0.14 (0.05) & 0.21 (0.16) &   0.84 (0.22) & 0.08 (0.07) & 0.41 (0.12) \\
Wine  & -3.94 (0.52) & -3.94 (0.05) &  -0.85 (0.06)&  0.32 (0.08) & -0.88 (0.04) & -0.86 (0.04) \\
Yacht & -0.34 (0.02) & -0.40 (0.03) & 0.12 (0.27) &  0.99 (0.09) & -0.25 (0.11) & -0.75 (0.06) \\
Kin8mn & -1.12 (0.01) & -1.29 (0.02)  & -1.29 (0.00) &  -0.19 (0.04) & -1.14 (0.02) & -1.14 (0.02)\\
Naval& -9.74 (0.02) & -8.87 (0.22)  & -9.92 (0.00) & -5.40 (0.10) & -9.84 (0.01) & -9.82 (0.01) \\
Power & 2.84 (0.03) & 2.33 (0.03) & 2.41 (0.02)& 3.11 (0.15)  & 2.57 (0.05) & 2.59 (0.05) \\
\end{tabular}
\caption{A comparison of kernel learning approaches for UCI benchmarks. NLPD ($\pm$ standard error of mean) evaluated on average of 10 splits with 90$\%$  of the data used for training.}
\label{tab:nlpd}
\end{table*}



\begin{table*}[t]
\centering

\begin{tabular}{l|c|c|c|c|c|c}
Dataset & RBF & TS & NKN  & AHGP & KITT (E + D) & KITT (E) \\
\hline
Boston & 3.06 (0.21) & 3.12 (0.29) & 2.51 (0.15) & 2.73 (0.38) & 3.14 (0.27) & 2.56 (0.13) \\
Concrete & 4.89 (0.17) & 3.83 (0.18) & 3.69 (0.24) & 3.45 (0.45) & 3.75 (0.18)  &  3.79 (0.2)  \\
Energy & 0.43 (0.02) & 0.28 (0.01) & 0.25 (0.02) &  0.51 (0.07) & 0.26 (0.012) & 0.28 (0.012) \\
Wine  & 0.65 (0.01) & 0.55 (0.01) & 0.52 (0.01) &  0.58 (0.04) & 0.546 (0.008) & 0.545 (0.0074)\\
Yacht  & 0.22 (0.02) & 0.22 (0.01) & 0.31 (0.06) &  0.46 (0.27) & 0.227 (0.024) & 0.187 (0.014)\\
Kin8mn & 0.08 (9e-04) & 0.08 (0.00)  & 0.07 (0.00) &  0.19 (0.01) & 0.078 (0.001) & 0.078 (0.0011)\\
Naval & 1e-05 (5e-07) & 0.00 (0.00) & 0.00 (0.00) &  0.00 (0.00) & 1.58e-5 (5e-7) & 1.6e-5 (3.5e-7) \\
Power & 4.13 (0.12) & 3.33 (0.16) & 2.68  (0.07)&  4.23 (0.24) & 3.33 (0.17) & 3.36 (0.16) \\
\end{tabular}
\caption{A comparison of kernel learning approaches for UCI benchmarks. RMSE ($\pm$ standard error of mean) evaluated on average of 10 splits with 90$\%$  of the data used for training. The acronyms used here are defined in Table \ref{tab:comp}.}
\label{tab:rmse}
\end{table*}




\begin{table*}[t]
\centering
\begin{tabular}{l|l|l|l}
Dataset & Random  & KITT Top 1 & KITT Top 3 \\
\hline
Boston & 3.48 (0.56) & 3.24 (0.18) & 2.56 (0.13) \\
Concrete & 5.75 (0.91) & 4.03 (0.21) & 3.79 (0.2) \\
Energy  & 1.43 (0.93) &  0.386 (0.012) &  0.28 (0.012) \\
Wine  &  0.66 (0.02) & 0.55 (0.0077) & 0.545 (0.0074) \\
Yacht  & 2.15 (0.95) & 0.209 (0.016) &  0.187 (0.014)  \\
Kin8mn  & 0.14 (0.02) & 0.08 (9e-04) & 0.078 (0.0011)  \\
Naval  & 0.01 (0.01) &  1.68e-05 (3.7e-07) & 1.6e-05 (3.5e-07)  \\
Power  & 5.43 (1.26) & 3.44 (0.14) & 3.36 (0.16)  \\
\end{tabular}
\caption{Predictive performance (RMSE) for three different model averaging strategies. }
\label{tab:strats}
\end{table*}



\begin{table*}[t]
\centering

\begin{tabular}{l|l|l|l|l}
Dataset & $N$ & $d$ & TS & KITT (E) \\
\hline
Boston & 506 & 13 & 450.3 (32.6) & 0.023 (0.001) \\
Concrete & 1030 & 8 & 1701.2 (121.4) & 0.023 (0.001) \\
Energy & 768 & 8 & 2936.8 (364.1) & 0.023 (0.001)\\
Wine & 1599 & 11 & 2988.0 (409.5) & 0.022 (0.001)\\
Yacht & 308 & 6 & 352.9 (20.3) & 0.023 (0.002)\\
Kin8mn & 8192 & 8 & 3392.9 (104.7) & 0.022 (4e-04)\\
Naval &  11934 & 14 & 6194.2 (677.8) & 0.022 (4e-04)\\
Power & 9568 & 4 & 3174.1 (166.8) & 0.022 (4e-04)\\
\end{tabular}
\caption{Time taken (in seconds) to select a kernel, on an Intel i7-8700 12 CPU cores (3.20GHz, 32GB RAM) with one GPU (GTX 1070, 8GB RAM). We report means with 1 std. deviation estimated from 10 repetitions. For the Tree Search (TS) times, only a maximum of $N=2,000$ datapoints are used.}
\end{table*}


\begin{figure}[t]
    \centering
    \includegraphics[scale=0.5]{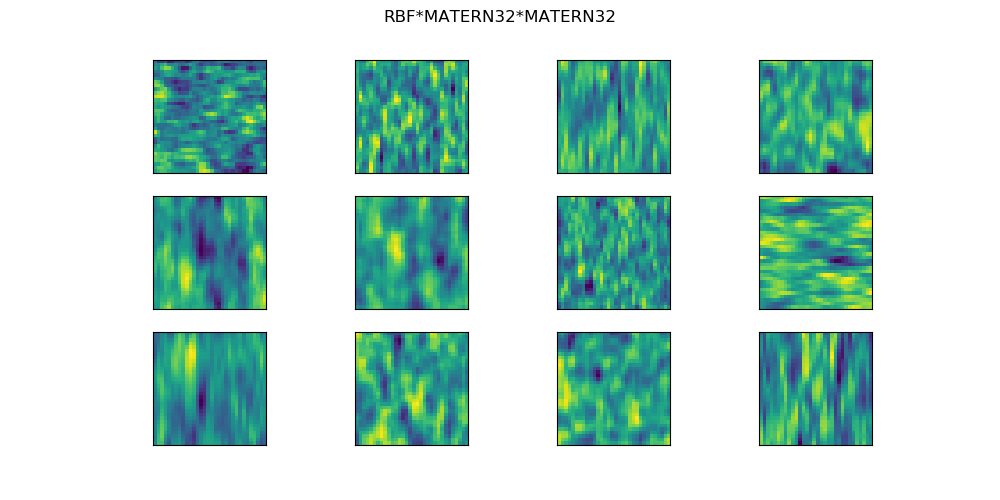}
    \includegraphics[scale=0.5]{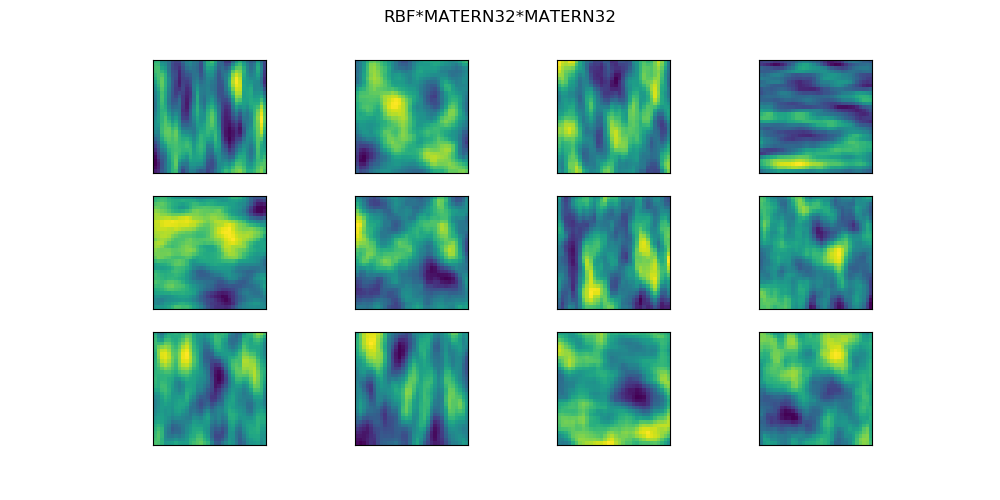}
    \caption{Random samples drawn from a triple product kernel across two input dimensions - before (top) and after (bottom) accounting for the induced shrinkage in lengthscale.}
    \label{fig:shrink}
\end{figure}

\begin{figure}[t]
    \centering
    \includegraphics[scale=0.4]{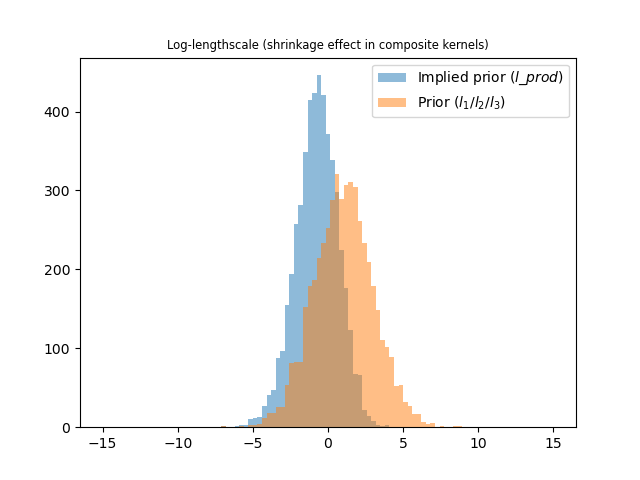}
    \includegraphics[scale=0.4]{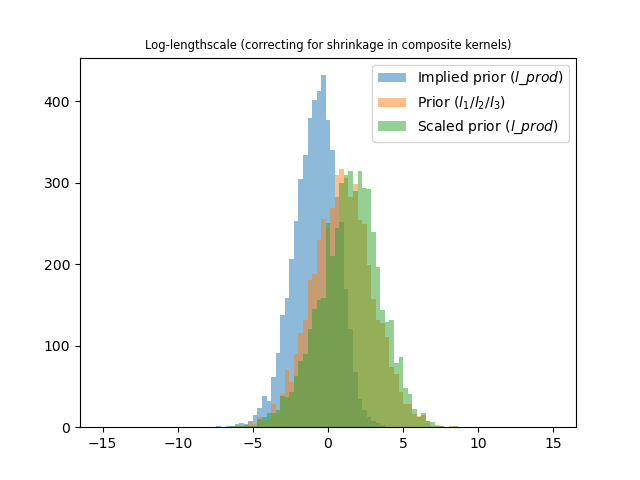}
    \caption{\textit{Left:} Histogram of samples from the lengthscale prior distribution and the implied prior distribution for a product of three RBF kernels. \textit{Right:} Additionally, samples (green) from the scaled prior ensuring that the range of lengthscales in the final product kernel has the same width as the component kernels.}
    \label{fig:shrink_dist}
\end{figure}

\section{Priors}





An important step in the construction of the training data  is defining a suitable set of priors for the hyperparameters for each kernel in the vocabulary. We select priors for each individual hyperparameter of the component kernels ensuring a wide support as discussed in the main text. For product kernels we learn a single variance hyperparameter, for instance, $\sigma^{2}_{f}(k_{1}k_{2}k_{3})$ but each additive term has its own variance hyperparameter. 
Since the priors are assigned on individual lengthscales in each component kernel, the prior on the implied lengthscale ends up having compressed support over a narrower range of shorter lengthscales (see figure \ref{fig:shrink}). Hence, the priors for the component lengthscale hyperparameters in product kernels are scaled to ensure a target prior over the implicit lengthscale. This is an important correction to make, because otherwise the network will learn to identify product kernels based upon their characteristically shorter lengthscales.

\textbf{Remark 1:} The product of two RBF kernels with identical lengthscales, $l_{1} = l_{2}$, yields another RBF kernel with lengthscale $l_{prod} = \dfrac{l_1}{\sqrt{2}}$. The product of three RBF kernels with identical lengthscales, $l_{1} = l_{2} = l_{3}$, yields another RBF kernel with lengthscale $l_{triple} = \dfrac{l_{1}l_{2}l_{3}}{\sqrt{l_{1}^{2} + l_{2}^{2} + l_{3}^{2}}}$.


\textbf{Remark 2:} If lengthscales $l_{1}, l_{2}, l_{3} \sim \mathcal{LN}(\mu,\sigma^{2})$ are independent log-normal random variables, then their product  $l_{1}l_{2}l_{3} \sim \mathcal{LN}(3\mu, 3\sigma^{2})$ and $\sqrt{(l_{2}l_{3})^{2} + (l_{1}l_{3})^{2} + (l_{1}l_{2})^{2}} \dot\sim \mathcal{LN}(0.5\mu_{z}, 0.25\sigma^{2}_{z})$ where,

\begin{align}
   \sigma^2_z &= \ln\!\left[ \frac{\sum e^{2\mu+\sigma^2}(e^{\sigma^2}-1)}{(\sum e^{\mu+\sigma^2/2})^2} + 1\right], \\
  \mu_z &= \ln\!\left[ \sum e^{\mu+\sigma^2/2} \right] - \frac{\sigma^2_z}{2} , 
\end{align}

is an approximation to the sum of log-normally distributed random variables \citep{asmussen2008asymptotics}.

Hence, $l_{triple} \dot\sim \mathcal{LN}(3\mu - 0.5\mu_{z}, 3\sigma^{2} + 0.25\sigma^{2}_{z} - 2\rho(\sqrt{3}\sigma)(\sqrt{0.25}\sigma_{z}))$ where $\rho$ is the correlation coefficient. Since we can deduce how the implied lengthscales in product kernels are approximately distributed, we can compute an approximate scaling factor for the priors in each of the component kernels such that the overall shrinkage is compensated for. This corection is shown in Figure \ref{fig:shrink_dist}.    The propagation of lengthscales will differ slightly for kernels whose spectral densities are non-Gaussian, however this is a relatively small impact.

\section{Experimental Details}

\subsection{Ground-Truth Experiments}

For each combination of test-size and dimensionality, we draw 300 GP samples from known primitive kernels and report test time classification accuracy. We report accuracy in terms of the fraction $(\%)$ of samples classified to the ground truth primitive kernel (i.e. correct classifications). We report means and standard errors over three runs per combination. Further, we report top-3 accuracy for the test size experiment and top-3/top-5 accuracy for the test dimensions experiment where we fixed the size of test inputs to $N = 1600$ across dimensions. 

\subsection{ML-II Training}
For all experiments, the initial set of hyperparameter values was determined by selecting the highest marginal likelihood from $1,000$ random draws from the priors.

\section{Broader Impact}
Selecting a kernel for a dataset to be modelled with a GP is a long-standing challenge and research goal for GP researchers. Techniques to learn the functional form of a kernel are usually extremely computationally intensive and can deter GP practitioners from using them. The common approach is to use a reasonable default and focus ones efforts on scalable training once a kernel has been fixed. We believe that a pre-trained solution such as the one we propose in our paper which removes the burden of expensive kernel selection at experiment time may enjoy wide-spread adoption and popularity. A sufficiently large pre-trained network available for direct download and inference enables users to cross-check and compare their hand-crafted kernels with automated proposals. 

\clearpage
\bibliographystyle{plainnat}
\bibliography{kitt}